\pdfoutput=1

\documentclass[11pt]{article}

\usepackage[final]{acl}

\usepackage{times}
\usepackage{latexsym}

\usepackage[T1]{fontenc}

\usepackage[utf8]{inputenc}

\usepackage{microtype}

\usepackage{inconsolata}
\usepackage{amsmath}
\usepackage{graphicx} 
\usepackage{booktabs}
\usepackage{multirow}
\usepackage{hyperref}

\title{Long Context is Not Long at All:\\ A Prospector of Long-Dependency Data for Large Language Models}

\author{
 \textbf{Longze Chen}\textsuperscript{1,2}\footnotemark[1]\quad
 \textbf{Ziqiang Liu}\textsuperscript{1,2}\footnotemark[1]\quad
 \textbf{Wanwei He}\textsuperscript{1,2}\footnotemark[1]\quad
 \textbf{Yunshui Li}\textsuperscript{1,2}\quad
 \textbf{Run Luo}\textsuperscript{1,2}\quad
 \textbf{Min Yang}\textsuperscript{1}\footnotemark[2]
\\
 \textsuperscript{1}Shenzhen Institute of Advanced Technology, Chinese Academy of Sciences
\\
 \textsuperscript{2}University of Chinese Academy of Sciences
\\
 \texttt{\{lz.chen2, zq.liu4, ww.he, min.yang\}@siat.ac.cn}
}

\begin{document}
\maketitle

\renewcommand{\thefootnote}{\fnsymbol{footnote}}
\footnotetext[1]{Equal contribution.}
\footnotetext[2]{Corresponding author.}

\renewcommand{\thefootnote}{\arabic{footnote}}

\begin{abstract}
Long-context modeling capabilities are important for large language models (LLMs) in various applications. However, directly training LLMs with long context windows is insufficient to enhance this capability since some training samples do not exhibit strong semantic dependencies across long contexts.
In this study, we propose a data mining framework \textbf{ProLong}\footnote{\ \  \url{https://github.com/October2001/ProLong}} that can assign each training sample with a long dependency score, which can be used to rank and filter samples that are more advantageous for enhancing long-context modeling abilities in LLM training. Specifically, we first use delta perplexity scores to measure the \textit{Dependency Strength} between text segments in a given document. Then we refine this metric based on the \textit{Dependency Distance} of these segments to incorporate spatial relationships across long-contexts. Final results are calibrated with a \textit{Dependency Specificity} metric to prevent trivial dependencies introduced by repetitive patterns. Moreover, a random sampling approach is proposed to optimize the computational efficiency of ProLong. Comprehensive experiments on multiple benchmarks indicate that ProLong effectively identifies documents that carry long dependencies and LLMs trained on these documents exhibit significantly enhanced long-context modeling capabilities.
\end{abstract}

\begin{figure}[t]
    \centering
    \includegraphics[width=0.9\columnwidth]{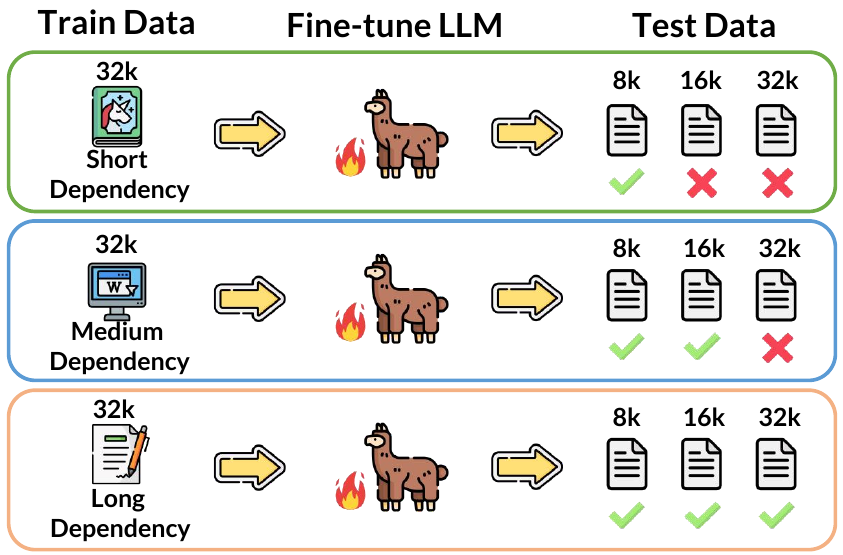}
    \caption{Samples that carry longer dependencies better enhances LLMs' long-context modeling capabilities, even with a fixed training context window of 32k.}
    \label{fig:overview}
\end{figure}

\section{Introduction}
Large language models (LLMs) are widely used in many natural language processing (NLP) tasks~\cite{brown2020language}. These tasks often require dealing with long text inputs~\cite{bai2023longbench, zhang2023marathon}, such as lengthy documents~\cite{zhou2022fine}, long conversation histories in chatbots~\cite{zhong2023memorybank} or large codebases~\cite{guo2023longcoder}. Therefore enhancing LLMs to model long-context inputs is a prominent desiderata.

There are primarily two categories of approaches to expand the context window of an LLM. 
The first category fine-tunes LLMs with longer context windows \cite{chen2023extending}, while the second category adjusts the LLM's positional encoding or attention mechanism to accommodate larger position indices without further re-training \cite{press2021train, ntkaware}.
However, non-training methods often produce results inferior to those of fine-tuned LLMs \cite{xiong2023effective}, which model long-context inputs more effectively and generally achieve lower perplexity scores. 

Although reported to be feasible, simply fine-tuning LLMs with naively sampled long corpora does not ensure improved long context modeling capabilities \cite{longrun2023}. 
Some of these fine-tuned LLMs may still struggle to effectively process and utilize information from long input contexts even if they obtain a decently low perplexity score \cite{pal2023giraffe}. 
This can lead to low performance in various downstream applications, even in some basic synthetic retrieval tasks \cite{liu2023lost}.
Nevertheless, few approaches try to tackle this long-context modeling issue from a data-centric perspective.

As revealed by \citet{fu2023longdata}, the quality of fine-tuning data plays a critical role in enhancing the long-context modeling capabilities of LLMs~\cite{li2023one}.
Besides, \citet{longrun2023} also report that high-quality corpora significantly outperform other factors in boosting long-context performance.
Upon further exploration, we recognize that high-quality long-text data is characterized by the presence of \textbf{long-range dependencies}.
The importance of encapsulating long-range dependencies in LLMs is also underscored by \citet{borgeaud2022improving}, which elucidates the benefits of integrating global dependencies into retrieval-augmented language models.

However, such strong semantic dependencies are rare in typical training samples \cite{longrun2023} and diminish as the distance between segments increases \cite{staniszewski2023structured}.
Even with identical sequence lengths, different samples may exhibit varying dependency density.
Specifically, certain long training samples often comprise concatenated short documents that are randomly selected and do not have any semantic dependencies. Moreover, even for inherently long documents, like novels, most tokens depend only to a brief span of preceding context.
This phenomenon can be simply concluded as ``\textbf{long context is not long at all}'', leading to challenges in model learning (Figure \ref{fig:overview}).
Therefore, we argue that explicitly incorporating long-dependency data into the fine-tuning process can facilitate long context modeling.

In this paper, we propose a novel framework (called ProLong) to mine long-dependency data.
ProLong assigns a long-dependency score to each document, which serves as an indicator of the dependency density across long contexts.
Documents with higher scores are deemed more advantageous for boosting long context modeling, therefore we can use these scores to rank and filter high quality corpus for LLM fine-tuning.
Concretely, ProLong first partitions each document into fixed-length segments and evaluates dependency relationships between each segment pair from three perspectives:
(i) \textit{dependency strength} quantifies the difference in perplexities of a given segment when conditioned with or without its preceding segments. This metric measures the contribution of the preceding segment to the current one;
(ii) \textit{dependency distance} measures the positional gap and spatial relationship between two text segments;
and (iii) \textit{dependency specificity} employs entropy to ensure a non-uniform distribution of dependency strengths across all preceding segments, mitigating trivial dependencies introduced by repetitive patterns.
A dependency score is assigned to each segment pair by combining the above three perspectives and a final long-dependency score for the entire document is computed by accumulating dependency scores of all segment pairs.

Further, we adopt various strategies to optimize the computational efficiency of ProLong, including sampling among segments, evaluating perplexity scores with small models and curating test sets for rapid validation. 
Experiments on multiple benchmarks indicate that ProLong effectively identifies documents that carry long dependencies and LLMs trained on these documents exhibit significantly enhanced long-context modeling capabilities.
Our contributions are summarized as follows:

\textbf{1.} To the best of our knowledge, this is the first study to explore the relationship between dependency density and the quality of long-text data.

\textbf{2.} We propose ProLong, a data mining framework for identifying long-dependency data. With ProLong, significant performance boosts are observed using only 50\% of fine-tuning data.
  
\textbf{3.} We provide an in-depth analysis of ProLong's components, optimizing computational efficiency and making it practical for large-scale corpora.

\textbf{4.} We develop two models, ProLong-7b/13b, using training samples derived by the ProLong framework.
Experiments show that our models outperform equal-sized competitors on both language modeling and real long-context tasks.

\section{Related Work}
\paragraph{Long-context LLMs.} Large Language Models (LLMs), such as Llama \cite{touvron2023llama} with a context size of 2048 and Llama2 \cite{touvron2023llama2} with 4096, are typically pre-trained using a pre-determined context size. However, training LLMs from scratch with extended contexts is extremely time-consuming and labor-intensive. Therefore, recent research endeavors to extend the limited context length of these models via fine-tuning.
Fine-tuning methods such as Positional Interpolation \cite{chen2023extending}, "NTK-aware" interpolation \cite{ntkaware}, Giraffe \cite{pal2023giraffe} and YaRN \cite{peng2023yarn} modify rotary position encoding (RoPE) \cite{su2024roformer} and then fine-tune the model with a small number of training steps to expand the context window. 
Despite this, full fine-tuning is still computationally expensive. Consequently, researchers have pivoted towards exploring methods to reduce training costs. For instance, LongLora \cite{chen2023longlora} proposes $\text{S}^2\text{-Attn}$ and utilizes LoRA \cite{hu2021lora} for low-cost and efficient training; Soaring \cite{zhang2024soaring} introduces Activation Beacon, achieving a dramatic extension of LLMs context at low training costs.
Nonetheless, these methods have not taken into account the quality of the long-context data used during fine-tuning stage. In contrast, our approach emphasizes the high-quality long-context training data, facilitating the efficient extension of LLMs' context with a limited quantity of data.

\paragraph{Constructing Long-context Training Data.} 
Long-context data is not simply the arbitrary concatenation of unrelated short texts. The construction of high-quality or specialized forms of long-context data also holds significant research value. 
\citet{longrun2023} argues against squandering attention on randomly connected long texts and instead propose considering strategies for obtaining meaningful long-context data. 
SPLICE \cite{staniszewski2023structured} introduces structured packing for long-context, constructing training examples by assembling multiple similar documents selected through a retrieval method. 
Ziya-Reader \cite{junqing2023never} constructs a tailored Multi-doc QA task that requires concentration on different positions in contexts to address the "lost in the middle" \cite{liu2023lost} problem. 
\citet{yu2023paraphrasing} delves into the required data format for training long-context models, claiming that the use of "W-shaped data" is essential to tackle the "lost in the middle" problem. 
Upon further exploration, we identify that high-quality long-text data is characterized by the presence of long-range dependencies. Our method focuses on selecting these long-dependency data from a substantial corpus of long-context data.

\begin{figure*}[t]
  \centering 
  \includegraphics[width=1.0\textwidth]{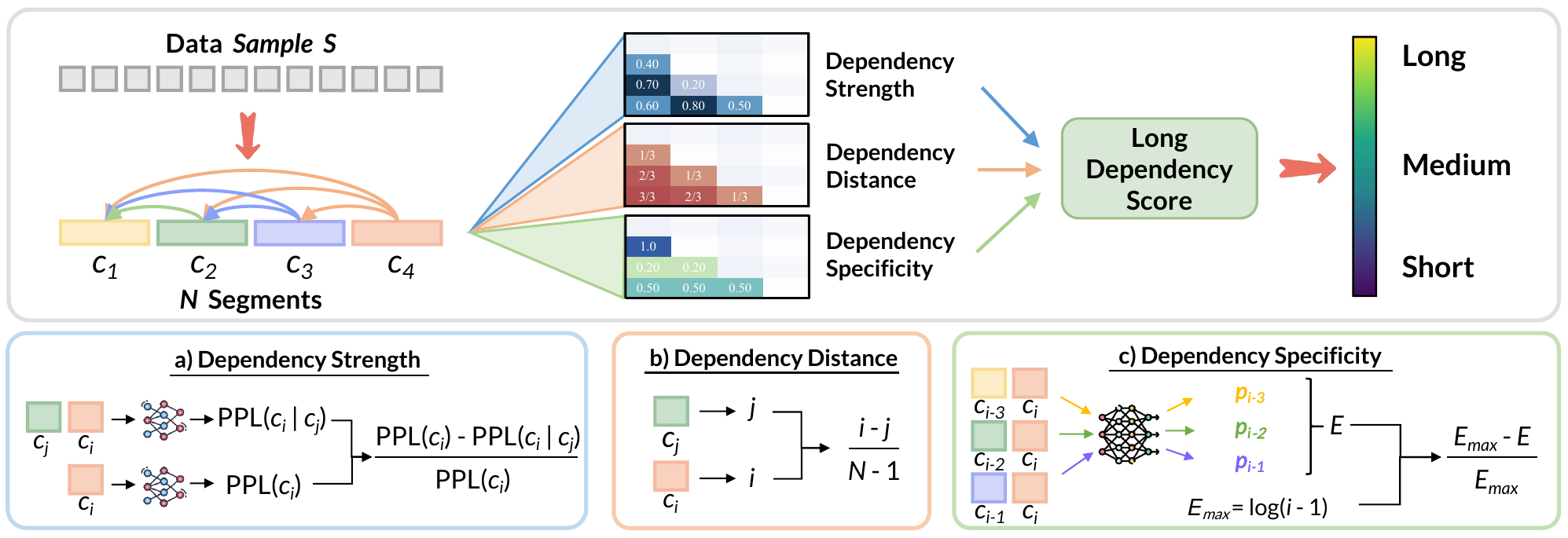}
  \caption{ProLong first segments a training sample $S$ into $N$ equal-length portions, then computing three key metrics: (a) dependency strength, (b) dependency distance, and (c) dependency specificity. These metrics are integrated via Eq.\ref{eq:lds} to derive the long dependency score for $S$.}
  \label{fig:architecture}
\end{figure*}

\section{ProLong Framework}
Figure \ref{fig:architecture} presents the overall framework of ProLong, which assigns a Long Dependency Score (LDS) for a given data sample $S$ ($S$ could be an intact document or a concatenated training sample).
Specifically, we first divide $S$ into $N$ segments of equal length $c_1, ..., c_N$.
Then for each segment pair $(c_j, c_i), j<i$, three scores are calculated: 
1. The Dependency Strength (DST) score evaluates how much semantic dependencies are there between $c_i$ and $c_j$;
2. The Dependency Distance (DDI) score demonstrates the distance between $c_i$ and $c_j$;
3. The Dependency Specificity (DSP) score penalizes the erroneous gains introduced by similar patterns in each segments. 
The final LDS for $S$ is obtained by merging the above three sets of scores.

\subsection{Dependency Strength}
For a given segment pair $(c_j, c_i), j<i$, the dependency between $c_i$ and $c_j$ can be evaluated by the delta perplexity score of $c_i$ when $c_j$ is given or not in the context. Specifically, if there are strong semantic dependencies between $c_i$ and $c_j$, then it is expected that the perplexity score of $c_i$ will be dramatically reduced if $c_j$ is presented in the context. Therefore, we can select a language model and use the difference of perplexity scores to quantify the dependency strength $\mathrm{DST}_{i,j}$ between $c_i$ and $c_j$:
\begin{equation}\label{DST}
\mathrm{DST}_{i, j} = \frac{\mathrm{PPL}\left(c_i\right) - \mathrm{PPL}\left(c_i \mid c_j\right)}{\mathrm{PPL}\left(c_i\right)}
\end{equation}
where $\mathrm{PPL}\left(c_i\right)$ is the perplexity score of $c_i$ without any context, and $\mathrm{PPL}\left(c_i \mid c_j\right)$ is the conditional perplexity score of $c_i$ when $c_j$ is provided in the input context. In this study, we use a small-sized and fixed model to calculate perplexity scores.

\subsection{Dependency Distance}
Another factor to consider in long dependency relationships is the distance between $c_i$ and $c_j$. 
Specifically, distant segments are more important for learning long-range dependencies.
Therefore for each pair $(c_j, c_i), j<i$, we consider a dependency distance score $\mathrm{DDI}_{i, j}$ to measure the positional gap:
\begin{equation}\label{DDI}
\mathrm{DDI}_{i, j} = \frac{i - j}{N - 1}
\end{equation}
where $N$ is the total segment count.

\subsection{Dependency Specificity}
Our early experiments observe that some documents in the training corpus contain repeated text spans. Specifically, in the extremest case, the entire document contain only one unique token. For a given segment pair $(c_j, c_i), j<i$ from these documents, $c_i$ and $c_j$ is nearly the same and $c_i$ will receive nearly perfect conditional perplexity score if $c_j$ is presented in the context. In this case, we will achieve an extremely high dependency strength score $\mathrm{DST}_{i,j}$ between $c_i$ and $c_j$.

However, such trivial dependencies brought by repeated text spans are usually harmful to the training of LLMs \cite{lee2021deduplicating}.
Therefore, we introduce the concept of dependency specificity (DSP) to mitigate this issue. 
Specifically, for a given segment $c_i$, we can calculate the reduction of perplexity scores $\Delta \mathrm{PPL}_{i,j} = \mathrm{PPL}(c_i) - \mathrm{PPL}(c_i|c_j)$ when conditioning $c_i$ on each of its preceding segments $c_j$, $(j=1, ..., i-1)$. A probability distribution $P(c_i) = (p_1, ..., p_{i-1})$ can then be obtained by applying Softmax on $\Delta \mathrm{PPL}_{i,j}$:
\begin{equation}\label{p}
p_j = \frac{\exp\left(\Delta\mathrm{PPL}_{i, j}\right)}{\sum_{k=1}^{i-1}\exp\left(\Delta\mathrm{PPL}_{i, k}\right)}
\end{equation}
$P(c_i)$ denotes the dependency distribution of $c_i$ on all preceding segments.
A uniform $P(c_i)$ indicates that $c_i$ relies almost equally on each preceding segment $c_j$ $(j=1, ..., i-1)$, suggesting that $c_j$ are nearly identical. Consequently, the entropy of $P(c_i)$ can serve to adjust the DST score derived from Eq.\ref{DST}, addressing the issue of repeated spans.

In this study, we define the dependency specificity $\mathrm{DSP}_i$ score for a given segement $c_i$ as
\begin{equation}\label{DSP}
\mathrm{DSP}_i = \frac{E_{max} - E}{E_{max}}
\end{equation}
where $E_{max} = \log\left(i - 1\right)$ is the entropy of a $i-1$ dimensional uniform distribution. And $E$ is the entropy of $P(c_i)$, i.e., $E = -\sum_{j=1}^{i-1}p_j \log\left(p_j\right)$. A higher $\mathrm{DSP}_i$ value suggests that the segment $c_i$ does not depend equally on all its preceding segments, indicating a lower likelihood of these segments being repeated spans.

\subsection{Long Dependency Scores}
We define a long dependency score between each segment pair $(c_j, c_i), j<i$ as:
\begin{equation}
\label{eq:lds_pair}
\begin{aligned}
\mathrm{LDS}_{i,j} = (\alpha \mathrm{DST}_{i, j} + \beta \mathrm{DDI}_{i, j}) \cdot \mathrm{DSP}_i
\end{aligned}
\end{equation}
where $\alpha, \beta$ are hyper-parameters that control the impact of $\mathrm{DST}_{i, j}$ and $\mathrm{DDI}_{i, j}$, respectively. 
Note that the formulation of Eq.\ref{eq:lds_pair} is essentially an adjustment of the $\mathrm{DST}_{i, j}$ score.
Concretely, if $c_i$ and $c_j$ exhibits strong semantic dependency over extended contexts, it is expected that $c_i$ and $c_j$ are spatially far (i.e., with large $\mathrm{DDI}_{i,j}$) and achieve a high $\mathrm{DST}_{i,j}$, and $\mathrm{DST}_{i,j}$ attributed to genuine semantic connections rather than repeated spans (as indicated by high $\mathrm{DSP}_i$).

Totally, we define the long-dependency score for the entire sample $S$ by accumulating $\mathrm{LDS}_{i,j}$:
\begin{equation}
\label{eq:lds}
\begin{aligned}
\mathrm{LDS} = \sum_{i = 1}^N \sum_{j = 1}^{i - 1} \mathrm{LDS}_{i,j} \cdot I_{i, j}
\end{aligned}
\end{equation}
\begin{equation}\label{I}
I_{i, j} =
\begin{cases} 
1,  & \text{if }\mathrm{DST}_{i,j}>\tau \\
0, & \text{otherwise}
\end{cases}
\end{equation}
where $\tau$ is a threshold for $\mathrm{DST}_{i,j}$ to filter out noises in the perplexity calculation.

\subsection{Enhancing Computational Efficiency}
\label{sec: Enhancing Computational Efficiency}
The time complexity for calculating LDS in Eq.\ref{eq:lds} is $O(N^2)$, which is impractical when handling massive candidate samples.
To optimize the computational efficiency of LDS, we employ a method of random sampling.
First, we randomly sample $T$ index pairs: $\mathcal{D} = \{(x_1, y_1), ..., (x_T, y_T)\}$, where $x_t \in [1, N]$, $y_t \in [1, N]$ and $x_i < y_i$ for $t=1,...,T$. Each element $(x_t, y_t) \in \mathcal{D}$ corresponds to a segment pair $(c_{x_t}, c_{y_t})$. Then a sampled LDS is calculated as:
\begin{equation}
\label{Long Dependence Score Sample}
\begin{aligned}
\mathrm{LDS_\textit{sp}} = \sum_{(x_t, y_t) \in \mathcal{D}} \mathrm{LDS}_{x_t, y_t} \cdot I_{x_t, y_t}
\end{aligned}
\end{equation}
The computational of $\mathrm{LDS}_{sp}$ is substantially decreased, lowering the time complexity to $O(T)$.

\begin{table*}[!htbp]
\centering 
\resizebox{1.0\textwidth}{!}{
\begin{tabular}{lccc|ccc|ccc|ccc}
\toprule
& \multicolumn{3}{c}{\textbf{KV Retrieval (140 Pairs)}} & \multicolumn{3}{c}{\textbf{KV Retrieval (300 Pairs)}} & \multicolumn{3}{c}{\textbf{MQA (20 Documents)}} & \multicolumn{3}{c}{\textbf{MQA (30 Documents)}} \\
\cmidrule(lr){2-4}\cmidrule(lr){5-7}\cmidrule(lr){8-10}\cmidrule(lr){11-13} 
& \textbf{Full} & \textbf{Rand} & \textbf{ProLong} & \textbf{Full} & \textbf{Rand} & \textbf{ProLong} & \textbf{Full} & \textbf{Rand} & \textbf{ProLong} & \textbf{Full} & \textbf{Rand} & \textbf{ProLong} \\
\textbf{Data Percentage} & (100\%) & (50\%) & (50\%) & (100\%) & (50\%) & (50\%) & (100\%) & (50\%) & (50\%) & (100\%) & (50\%) & (50\%) \\
\cmidrule(lr){1-1}\cmidrule(lr){2-4}\cmidrule(lr){5-7}\cmidrule(lr){8-10}\cmidrule(lr){11-13}
Llama2-7b & 77.4 & 78.0 & \textbf{93.4} & 59.5 & 52.6 & \textbf{86.0} & 43.9 & 43.8 & \textbf{45.2} & 42.6 & 42.8 & \textbf{44.4} \\
Llama2-13b & 95.3 & 94.5 & \textbf{95.4} & 84.1 & 82.2 & \textbf{84.1} & 46.2 & 48.4 & \textbf{49.1} & 43.0 & 45.9 & \textbf{46.1} \\
\midrule[0.5pt]
\midrule[0.5pt]
& \multicolumn{3}{c}{\textbf{HotpotQA}} & \multicolumn{3}{c}{\textbf{2WikiMultihopQA}} & \multicolumn{3}{c}{\textbf{MuSiQue}} & \multicolumn{3}{c}{\textbf{GovReport}} \\
\cmidrule(lr){2-4}\cmidrule(lr){5-7}\cmidrule(lr){8-10}\cmidrule(lr){11-13} 
& \textbf{Full} & \textbf{Rand} & \textbf{ProLong} & \textbf{Full} & \textbf{Rand} & \textbf{ProLong} & \textbf{Full} & \textbf{Rand} & \textbf{ProLong} & \textbf{Full} & \textbf{Rand} & \textbf{ProLong} \\
\textbf{Data Percentage} & (100\%) & (50\%) & (50\%) & (100\%) & (50\%) & (50\%) & (100\%) & (50\%) & (50\%) & (100\%) & (50\%) & (50\%) \\
\cmidrule(lr){1-1}\cmidrule(lr){2-4}\cmidrule(lr){5-7}\cmidrule(lr){8-10}\cmidrule(lr){11-13}
Llama2-7b & 44.4 & 42.4 & \textbf{44.9} & \textbf{34.4} & 33.6 & 34.2 & \textbf{21.9} & 19.9 & 21.3 & 30.2 & 29.4 & \textbf{31.0} \\
Llama2-13b & \textbf{49.4} & 48.2 & 48.6 & 38.5 & 38.2 & \textbf{39.7}& 19.1 & 17.0 & \textbf{19.8} & 32.0 & 32.1 & \textbf{32.5} \\
\midrule[0.5pt]
\midrule[0.5pt]
& \multicolumn{3}{c}{\textbf{Qasper}} & \multicolumn{3}{c}{\textbf{SAMSum}} & \multicolumn{3}{c}{\textbf{LCC}} & \multicolumn{3}{c}{\textbf{RepoBench-P}} \\
\cmidrule(lr){2-4}\cmidrule(lr){5-7}\cmidrule(lr){8-10}\cmidrule(lr){11-13} 
& \textbf{Full} & \textbf{Rand} & \textbf{ProLong} & \textbf{Full} & \textbf{Rand} & \textbf{ProLong} & \textbf{Full} & \textbf{Rand} & \textbf{ProLong} & \textbf{Full} & \textbf{Rand} & \textbf{ProLong} \\
\textbf{Data Percentage} & (100\%) & (50\%) & (50\%) & (100\%) & (50\%) & (50\%) & (100\%) & (50\%) & (50\%) & (100\%) & (50\%) & (50\%) \\
\cmidrule(lr){1-1}\cmidrule(lr){2-4}\cmidrule(lr){5-7}\cmidrule(lr){8-10}\cmidrule(lr){11-13}
Llama2-7b & \textbf{29.7} & 27.6 & 28.3 & 42.7 & 42.4 & \textbf{43.2} & 64.9 & 64.4 & \textbf{65.2} & 58.5 & 59.2 & \textbf{60.5} \\
Llama2-13b & 38.4 & 36.9 & \textbf{38.9} & 43.1 & 43.2 & \textbf{44.2} & 67.5 & 66.9 & \textbf{67.7} & \textbf{61.1} & 60.5 & 60.8 \\
\bottomrule
\end{tabular}}
\caption{Results of ProLong on Llama-2-7B, Llama-2-13B. Full denote full dataset, and otherwise we select 50\% of the data with random selection (Rand). \textbf{Bold} numbers denotes the best performing selected subset.}
\label{tab:main results}
\end{table*}

\section{Experimental Setup}
\subsection{Training Datasets} 
Our training datasets consist of three parts: pre-train dataset, English book dataset and code dataset.
The pre-training dataset is sampled from RedPajama \cite{together2023redpajama} to mitigate catastrophic forgetting induced by extended training. The pre-training dataset part keeps the same across all experiments.
We choose English books and code data primarily for extending training because they naturally possess sufficiently long lengths. Note that the length of each document in all datasets exceeds 32k.
Detailed training data information is provided in Appendix \ref{sec: train data}.

\subsection{Task Formulation}
Assume that there are multiple models based on different training datasets and methods.
We comprehensively evaluate the long context capability of the models, including three tasks as follows:
\paragraph{Language Modeling Tasks.}
Language modeling task is the most fundamental requirement for LLMs, typically measured by perplexity (PPL) on the test text data. We follow \citet{chen2023longlora} to select 128 documents that are randomly sampled from the total proof-pile \cite{proofpile} test split. For each document, it has at least 32768 tokens. 
All PPL results are calculated using the sliding window method \cite{press2021train} with stride $S = 1024$.

\paragraph{Synthetic Long Context Tasks.}
The synthetic long context tasks are divided into two sub-tasks: multi-document question answering (MQA) and key-value retrieval.
Following the methodology of \citet{liu2023lost}, we make controlled changes to the input context size and the position of the relevant information within this context in both sub-tasks.
In the MQA task, the total number of documents is either 20 or 30, corresponding to text lengths of 4k or 6k tokens respectively.
For the key-value retrieval task, we use a total of 140 or 300 key-value pairs, corresponding to total token counts of 8k or 16k respectively. The goal of both task is to evaluate the model's ability to accurately identify unique relevant information from a large volume of irrelevant documents (or KV pairs).

\paragraph{Real-World Long Context Tasks.}
The current mainstream real-world long context tasks are derived from LongBench \cite{bai2023longbench}, primarily including single-document question answering, multi-document question answering, summarization, code completion, with the average length of most tasks ranging from 5k to 15k.
We use the metrics and scripts provided along with the benchmark for evaluation.

\subsection{Data Selection} 
\label{sec: data sel}
In the experiments, we truncate the length $M$ of all data to $32768$ and set the segment length $L$ to $128$, thereby dividing each data instance into $N=256$ segments.
We set the hyper-parameters $\alpha = \beta = 1$ in LDS.
To improve ProLong's computational efficiency, we replace the standard LDS with $\mathrm{LDS}_\textit{sp}$, utilizing a sampling size of $T=5000$.
We adopt a small model OPT-350m \cite{zhang2022opt} for calculating perplexity in LDS.
We rank the book and code data based on their LDS, independently retaining those documents with high LDS from each source. These retained subsets plus pre-training dataset are then integrated to construct our final training dataset. This method not only ensures the quality of our data but also enhances its diversity.
In the experiments, we primarily compare the results of utilizing the entire dataset (Full), randomly selecting a 50\% subset (Rand), and selecting the top-scoring 50\% subset (ProLong) for extending training. We list the average LDS statistics under different data selection strategies in Appendix \ref{sec: avg lds}.

\subsection{Training Details}
We extend the pre-trained Llama2-7b/13b \cite{touvron2023llama2} models to support context windows as large as 32768 tokens. We employ the NTK-aware \cite{ntkaware} method with base 160000 to rescale the position indices, without any additional modifications to the Llama2 model architectures. 
Other hyper-parameters for training is displayed in Appendix \ref{sec: hpara}.
For model comparison, we evaluate several mainstream LLMs with long context capability listed in Appendix \ref{sec: model baseline}.

\section{Experimental Results}
\subsection{ProLong Effectiveness}
As shown in Table \ref{tab:main results}, we compare the long context performance of the models on different data selection methods (i.e. Full v.s. Rand $50\%$ v.s. ProLong $50\%$).
The evaluated benchmarks includes KV retrieval, multi-document QA and typical tasks in LongBench.
We find that based on the same model (Llama2-7b or Llama2-13b), ProLong consistently outperforms random data selection across all the long context benchmarks. This result underscores the efficacy of ProLong in mining data with strong semantic dependencies across long contexts, which in turn enhances the long context capabilities of LLMs.
Interestingly, Table \ref{tab:main results} also suggests that training with $50\%$ of the data selected by ProLong often yields better results than training with the full dataset. This observation implies that most long contexts in the full dataset are not truly long at all. The principle of ``less is more'' surprisingly applies to certain specific long context tasks.

\subsection{Ablation Study on ProLong}
\label{ablation}

\begin{table}[tbp]\normalsize
\centering
\resizebox{1.0\columnwidth}{!}{\setlength{\tabcolsep}{1.5mm}{
\begin{tabular}{l|c|c}
\toprule
\multirow{2}{*}{\textbf{Model Setup}} & \multicolumn{2}{c}{\textbf{Multi-Document QA}}\\
\cmidrule(lr){2-3}
& \textbf{20 Documents} & \textbf{30 Documents} \\ 
\midrule[0.5pt]
\text{ProLong}-7b & \textbf{45.2} & \textbf{44.4} \\
\ \ \ \ w/o DSP     & 43.2 (-2.0)  & 43.1 (-1.3) \\
\ \ \ \ \ \ \ \ w/ DSP-add   & 43.3 (-1.9)  & 43.2 (-1.2) \\
\ \ \ \ w/o DDI    & 43.5 (-1.7)  & 43.8 (-0.6) \\
\ \ \ \ w/o DST    & 42.7 (-2.5)  & 42.8 (-1.6) \\
\bottomrule
\end{tabular}}}
\caption{Ablation study on the MQA task.}
\label{tab:ablation}
\end{table}

\begin{table}[tbp]
\centering 
\resizebox{1.0\columnwidth}{!}{\begin{tabular}{l|ccccc}
\toprule
\multirow{2}{*}{\textbf{Model}} & \multicolumn{5}{c}{\textbf{Evaluation Context Window Size}} \\
& 2048 & 4096 & 8192 & 16384 & 32768 \\
\midrule[0.5pt]
Llama2-7b & 3.19 & 2.91 & 2.88 & 4.22 & 18.72 \\
Code Llama-7b & 3.53 & 3.17 & 2.93 & 2.77 & 2.84 \\
Yarn-7b-64k & 3.29 & 2.99 & 2.79 & 2.66 & 2.60 \\
\textbf{ProLong-7b} & \textbf{3.02} & \textbf{2.76} & \textbf{2.58} & \textbf{2.45} & \textbf{2.38} \\
\midrule[0.5pt]
Llama2-13b & 3.03 & 2.77 & 2.71 & 3.58 & 14.69 \\
Code Llama-13b & 3.40 & 3.06 & 2.82 & 2.67 & 2.71 \\
Yarn-13b-64k & 3.10 & 2.82 & 2.63 & 2.51 & 2.43 \\
\textbf{ProLong-13b} & \textbf{2.89} & \textbf{2.64} & \textbf{2.47} & \textbf{2.35} & \textbf{2.28} \\
\bottomrule
\end{tabular}}
\caption{Sliding window perplexity~(stride S=1024) of 128 32k-length Proof-pile documents truncated to evaluation context window size.}
\label{tab:ppl}
\end{table}

\begin{table*}[t]\normalsize
\centering
\setlength{\tabcolsep}{1.5mm}{
\resizebox{1.0\textwidth}{!}{
\begin{tabular}{l|c@{\hspace{3mm}}c@{\hspace{3mm}}c@{\hspace{3mm}}c@{\hspace{3mm}}c@{\hspace{3mm}}|c|c@{\hspace{3mm}}c@{\hspace{3mm}}c@{\hspace{3mm}}c@{\hspace{3mm}}c@{\hspace{3mm}}c@{\hspace{3mm}}c@{\hspace{3mm}}|c@{\hspace{3mm}}}
\toprule
\multirow{2}{*}{\textbf{Model}} & \multicolumn{6}{c|}{\textbf{140 Key-Value Pairs}} & \multicolumn{8}{c}{\textbf{300 Key-Value Pairs}} \\ 
& $\text{p}_1$ & $\text{p}_{35}$ & $\text{p}_{70}$ & $\text{p}_{105}$ & $\text{p}_{140}$ & Avg. & $\text{p}_1$ & $\text{p}_{50}$ & $\text{p}_{100}$ & $\text{p}_{150}$ & $\text{p}_{200}$ & $\text{p}_{250}$ & $\text{p}_{300}$ & Avg. \\
\midrule[0.5pt]
GPT-4-32k & 98.2 & \textbf{98.2} & 93.6 & 84.0 & \textbf{100.0} & 94.8 & 99.0 & \textbf{94.0} & 52.2 & 37.8 & 20.8 & 20.0 & \textbf{99.8} & 60.5 \\
GPT-3.5-Turbo-16k & \textbf{100.0} & 97.0 & 72.6 & 91.8 & 99.8 & 92.2 & \textbf{100.0} & 85.0 & 75.8 & 50.0 & 29.4 & 55.2 & 99.4 & 70.7 \\
Yarn-7b-64k & 97.6 & 19.2 & 11.0 & 29.0 & 92.8 & 49.9 & 71.6 & 10.8 & 1.6 & 1.0 & 4.6 & 3.8 & 73.2& 23.8 \\
Yarn-13b-64k &74.0 & 23.6 & 49.0 & 12.4 & 89.2 & 49.6 & 72.2 & 9.2 & 3.6 & 1.2 & 8.4 & 7.0 & 86.4 & 26.9 \\
Vicuna-v1.5-13B-16k & 97.8 &79.2 & 84.6 & 78.6 & 28.2 & 73.7 & - & - & - & - & - & - & - & - \\
\midrule[0.5pt]
\textbf{ProLong-7b} & 99.6 & 95.2 & 91.8 & 91.6 & 88.8 & 93.4 & 98.0 & 61.8 & \textbf{82.4} & \textbf{88.2} & 86.4 & 93.4 & 92.0 & \textbf{86.0} \\
\textbf{ProLong-13b} & \textbf{100.0} & 96.2 & \textbf{96.6} & \textbf{94.8} & 89.4 & \textbf{95.4} & 99.8 & 86.0 & 73.6 & 59.4 & \textbf{88.4} & \textbf{94.2} & 87.6 & 84.1 \\
\bottomrule
\end{tabular}}}
\caption{Key-value retrieval performance on dictionaries of 140 and 300 key-value pairs. $p_i$ denotes the setting that the relevant information is at the i-th positions. ``-'' denotes that the result is not applicable due to exceeding context length after tokenization.}
\label{tab:kvr}
\end{table*}

\begin{table*}[t]
\centering 
\resizebox{1.0\textwidth}{!}{
\begin{tabular}{l|cc|cc|c|c|cc|c}
\toprule
\multirow{3}{*}{\textbf{Model}} & \multicolumn{2}{c|}{\textbf{Single-Doc QA}} & \multicolumn{2}{c|}{\textbf{Multi-Doc QA}} & \multicolumn{1}{c|}{\textbf{Summarization}} & \multicolumn{1}{c|}{\textbf{Few-shot Learning}} & \multicolumn{2}{c|}{\textbf{Code}} & \multicolumn{1}{c}{\textbf{Overall}} \\ 
\cmidrule(lr){2-3}\cmidrule(lr){4-5}\cmidrule(lr){6-6}\cmidrule(lr){7-7}\cmidrule(lr){8-9}\cmidrule(lr){10-10}  
& \multirow{2}{*}{\begin{tabular}[c]{@{}c@{}}Narrative\\ QA\end{tabular}} & \multirow{2}{*}{Qasper} & \multirow{2}{*}{\begin{tabular}[c]{@{}c@{}}Hotpot\\ QA\end{tabular}} & \multirow{2}{*}{\begin{tabular}[c]{@{}c@{}}2WikiMulti\\ hopQA\end{tabular}} & \multirow{2}{*}{\begin{tabular}[c]{@{}c@{}}Gov\\ Report\end{tabular}} & \multirow{2}{*}{SAMSum} & \multirow{2}{*}{LCC} & \multirow{2}{*}{Repobench-p} & \multirow{2}{*}{All} \\
& & & & & & & & & \\ 
\midrule[0.5pt]
GPT-3.5-Turbo-16k & 23.6 & 43.3 & 51.6 & 37.7 & 29.5 & 41.7 & 54.7 & 53.6 & 42.0 \\
\midrule[0.5pt]
Llama2-7b-chat-4k & 18.7 & 19.2 & 25.4 & 32.8 & 27.3 & 40.7 & 52.4 & 43.8 & 32.5 \\
LongChat-v1.5-7b-32k & 16.9 & 27.7 & 31.5 & 20.6 & 30.8 & 34.2 & 53.0 & 55.3 & 33.8 \\
Vincuna-v1.5-7b-16k & 19.4 & 26.1 & 25.3 & 20.8 & 27.9 & 40.8 & 51.0 & 43.5 & 31.9 \\
LongLora-7b-16k & 19.8 & 29.1 & 37.0 & 30.3 & \textbf{31.5} & 41.9 & 57.6 & 54.5 & 37.7 \\
Yarn-7b-64k & \textbf{25.0} & \textbf{30.8} & 39.5 & 30.3 & 26.2 & 42.6 & 64.6 & 59.4 & 39.8\\
\textbf{ProLong-7b} & 23.5 & 28.3 & \textbf{44.9} & \textbf{34.2} & 31.0 & \textbf{43.2} & \textbf{65.2} & \textbf{60.5} & \textbf{41.4} \\
\midrule[0.5pt]
Llama2-13b-chat-4k & 19.2 & 25.8 & 36.1 & 32.4 & 26.6 & 36.5 & 51.9 & 52.8 & 35.2 \\
Vicuna-v1.5-13b-16k & 18.9 & 29.9 & 38.1 & 36.0 & 27.9 & 27.8 & 44.1 & 45.6 & 33.5 \\
PI-Llama2-13b-16k & 19.2 & 33.3 & 44.9 & 34.8 & 27.9 & 27.9 & 62.5 & 51.1 & 37.7\\
Yarn-13b-64k & 21.0 & 27.6 & 47.2 & 38.0 & 19.5 & 43.2 & 65.2 & 57.9 & 40.0\\
\textbf{ProLong-13b} & \textbf{23.0} & \textbf{38.9} & \textbf{48.6} & \textbf{39.7} & \textbf{32.5} & \textbf{44.2} & \textbf{67.7} & \textbf{60.8} & \textbf{44.4} \\
\bottomrule
\end{tabular}}
\caption{Results (\%) on single-doc QA, multi-doc QA, summarization, few-shot learning and code tasks from LongBench dataset. ‘Overall’ is computed
by the macro-average over major task categories.}
\label{tab:longbench}
\end{table*}

As shown in Table \ref{tab:ablation}, we also conduct ablation studies on multi-document question answering task.
We first investigate the effect of the dependency specificity by removing it (denote as ``w/o DSP''), which is motivated to mitigate trivial dependencies introduced by repetitive patterns. The accuracy decreases in two different total document settings. 
Moreover, we observe a significant presence of repetitive patterns in the top-ranked data under ``w/o DSP'' setting, with examples provided in Appendix \ref{sec: repetitive patterns}. Thus, dependency specificity term indeed plays an crucial role on preventing the retrieval of meaningless repetitive patterns.
Next, we examine the effect of using DSP as a multiplier instead of an addend in Eq. \ref{eq:lds} (refer to Appendix \ref{sec: add dsp} for the formula with DSP as an added bias, denoted as ``w/ DSP-add''). The results indicate that using DSP as a bias is insufficient to mitigate the trivial dependencies introduced by repetitive patterns, and the performance of experiments is similar to that without DSP.
Finally, we conduct experiments to analyze the impact of removing the dependency distance (denote as ``w/o DDI''), or removing the dependency strength (denote as ``w/o DST'') to do ablation study. The results show that dependency strength, as the core aspect, contributes the most to the overall performance. Furthermore, removing DDI also leads to a decrease in performance.
In conclusion, each term in the LDS calculation has its unique advantages and is indispensable in our ProLong framework. DST, DDI, and DSP collectively make the final LDS a feasible and effective index for mining data with long dependencies amidst a sea of long texts.

Given the effectiveness of ProLong, we train models from Llama2~\cite{touvron2023llama2} on the top of ProLong 50\% (refer to Sec. \ref{sec: data sel} for setups in detail), resulting in \textbf{ProLong-7b/13b}.  

\subsection{Performance on Language Modeling}
In Table \ref{tab:ppl}, we compare our models ProLong-7b/13b with other baseline models in language modeling in various context window size, which is measured by perplexity. 
Firstly, ProLong-7b/13b exhibit a trend where perplexity decreases as the context window size increases, suggesting that they are capable of performing better language modeling with longer contexts.
Additionally, under any context window size, our models based on ProLong significantly outperform other baseline models, including Llama2. This demonstrates that ProLong-7b/13b exhibit superior language modeling capabilities in long contexts.

\begin{figure*}[t]
  \centering 
  \includegraphics[width=1.0\textwidth]{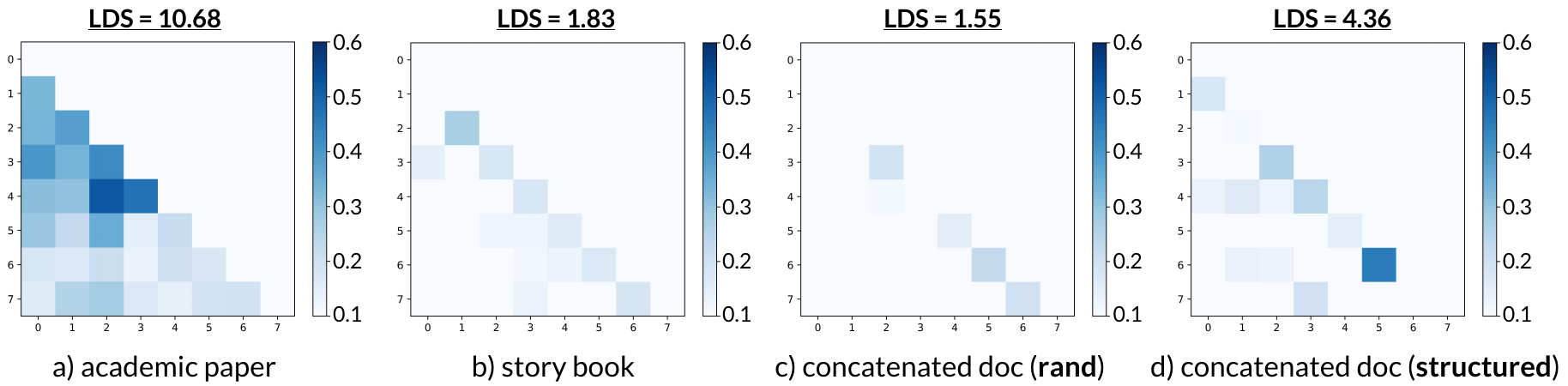}
  \caption{Visualization results of the dependency strength heat map for four data instances of equal length.}
  \label{fig: visualization}
\end{figure*}

\subsection{Performance on Key Value Retrieval}
\label{perf:kv-pair}
We claim that lower PPL does not fully encapsulate the ability to handle long text tasks effectively. Thus we also present the key value retrieval performance of ProLong-7b and ProLong-13b in Table \ref{tab:kvr}. 
Compared with other models based on Llama2, ProLong-7b/13b surpass them by a significant margin under the settings of 140 and 300 key-value pairs. Notably, ProLong-13b even outperforms the commercial model GPT-4-32k, especially in the more challenging setting of 300 key-value pairs.
One key gap lies in that in the middle position, while other models show a significant drop in performance, our models based on the ProLong framework maintain high performance. This suggests that the ``lost in the middle'' phenomenon, as described by \citet{liu2023lost}, is largely alleviated in our models.
We also conducted experiments on the ``needle in a haystack'' task, which bears similarity to the key-value retrieval task. Detailed experimental results can be found in Appendix \ref{sec: needle}. Consistent with the key-value retrieval task, the results demonstrate that the long context retrieval capability of ProLong-7b/13b experiences almost no degradation in the middle part of the context.

\subsection{Performance on LongBench}
\label{perf:longbench}
For comprehensive evaluation on the long context performance, we further use a real-world long context benchmarks \cite{bai2023longbench}.
As shown in Table \ref{tab:longbench}, we report the performance results (\%) of partial English and code tasks on the LongBench dataset.
Specifically, our ProLong-7b/13b achieves 41.1\%/44.4\% overall accuracy, which outperforms all the compared Llama2-based models by large margins. Besides, ProLong-13b surpasses the commercial model GPT-3.5-turbo-16k by an absolute gap 2.4\%. These results not only confirm the robust long-context capabilities of ProLong-7b/13b, but also suggest that the ProLong framework holds promise for effectively addressing real-world long-context tasks. We also conducted experiments on standard short-length benchmarks, which further demonstrates ProLong can enhance the long-context modeling ability without compromising other capabilities. Detailed experimental results can be found in Appendix~\ref{sec: short-bench}.

\section{Analysis}
\subsection{Balance between Accuracy and Efficiency}
Initial LDS requires $N^2$ times precise calculations. In practice, sampling $T\ll N^2$ times and using a small model for perplexity calculations provides a more efficient, albeit approximate, solution. 
In this section, we explore the balance between accuracy and efficiency.
We conduct experiments on the impact of different model sizes used for calculating perplexity (PPL) and different sampling granularity mentioned in Sec. \ref{sec: Enhancing Computational Efficiency}. 
To assess the accuracy of ProLong's capability in selecting long-dependency data, We construct a toy test set from various sources. Specifically, we manually construct 100 instances with strong long-dependency as positive examples and 100 instances with weak long-dependency as negative examples. Subsequently, we use ProLong to score and rank the test set, assessing the number (accuracy) of positive examples contained in the top 100 ranked data. 

We calculate the accuracy of ProLong in the test set to retrieve long-context documents and the number of documents processed per second as a measure of speed. The experimental results are as shown in Table \ref{tab: ProLong TestSet}, which reveals that smaller models and sampling sizes can achieve superior speed while sustaining comparable accuracy. In our experiments, we empirically choose OPT-350m and 5000 sampling size to perform ProLong, which may not necessarily yield the best results. Practically, it is essential to select an appropriate model and sampling size to strike a balance between accuracy and efficiency. 

\begin{table}[tbp]
\centering 
\resizebox{1.0\columnwidth}{!}{\begin{tabular}{lccc}
\toprule
\textbf{\# Samples} & \textbf{Model} & \textbf{Speed (documents/s)} & \textbf{Accuracy} \\
\midrule[0.5pt]
\multirow{3}{*}{5000} & OPT-350m & 0.16 & 89\% \\
& Qwen-1.8b & 0.04 & 89\%\\
& Llama2-7b & 0.01 & 96\% \\
\midrule[0.5pt]
\multirow{3}{*}{500} & OPT-350m & \textbf{1.11} & 87\%\\
& Qwen-1.8b & 0.32 & 87\%\\
& Llama2-7b & 0.08 & 96\% \\
\bottomrule
\end{tabular}}
\caption{Test result of ProLong by varying model sizes and sampling sizes.}
\label{tab: ProLong TestSet}
\end{table}

\subsection{Visualization Analysis}
To better illustrate the superiority of ProLong, we provide visualizations of dependency strength heat maps for various documents under the same settings.
Figure \ref{fig: visualization} displays various documents from the training set alongside their heat map visualization and LDS. LDS accurately capture the long-range dependencies within diverse documents. Specifically, academic papers typically showcase high LDS values, indicative of a more intricate heat map relationship and a pronounced presence of long-range dependencies. The same goes for and vice versa. Besides, we also find that the results of random concatenation of short texts (Figure \ref{fig: visualization}.c) are very similar to those in storybooks (Figure \ref{fig: visualization}.b), which demonstrates that even in naturally long texts, long-range dependencies may be minimal. 
Moreover, structured concatenated documents receive much higher LDS scores than randomly concatenated documents, corroborating part of the conclusions from the concurrent work, SPLICE~\cite{staniszewski2023structured}.

\section{Conclusion}
In this study, we propose ProLong, a novel and efficient approach for filtering long dependency data in language model extension training.
Benefiting from ProLong, we demonstrate that long-dependency data is the key to enhancing long-context modeling capability. Moreover, low-quality short-dependency data may even impair long-context modeling capability.
Extensive experiments on multiple long-context benchmark datasets demonstrate the effectiveness of our ProLong in enhancing the long-context modeling capability of large language models.
We hope ProLong can inspire researchers to spend more effort on data of more intrinsic long-dependency rather than long surface form.

\section*{Limitations}
The ProLong framework can be used to effectively identify documents that carry long dependencies and LLM trained on these documents exhibit significantly enhanced long-context modeling capabilities. However, to build strong LLMs, ProLong should be used in combination with other techniques. Here we list some of the limitations that are not considered when designing ProLong: 
(1) Diversity of Corpora. Our experiment only consider English books and code texts. In future work, we should extend our analysis to cover a wider range of long-context data, including different languages and document types.
(2) Model Considerations. In this study, our experiments are solely performed on Llama2-7B and Llama2-13B models. It is expected to obtain better performance with larger base-models.
(3) Data Mixture Percentage. We do not perform comprehensive exploration of the data mixture percentage, which is reported to be important to the final performance of LLM.

In future studies, we hope to make further explorations of the above listed limitations.

\section*{Ethics Statement}

Our study do not carry any ethical concerns. Specifically,  
Our training data are publicly available and designated for research purposes only. We inspect our dataset to ensure it does not contain any unethical content, private information and offensive topics.
Moreover, the base models we used are also publicly available for research purpose.

\bibliography{custom}

\clearpage
\appendix

\section{Training Datasets}
\subsection{Training Datasets Overview}
\label{sec: train data}
All training corpus we used in our experiments are given in Table \ref{tab:train-data}. 
The majority of the data consists of English books and code, all of which have lengths exceeding 32k. 
\begin{table}[h]
\centering
\resizebox{1.0\columnwidth}{!}{\begin{tabular}{llcc}
\toprule
\textbf{Type} & \textbf{Dataset} & \textbf{\# Instance} & \textbf{\# Len}  \\
\midrule[0.5pt]
\textbf{Pre-train Data} & RedPajama & 98k & 1.2k \\
\midrule[0.5pt]
\multirow{4}{*}{\textbf{English Book}} & arXiv & 200k & 54.7k \\
& Book3 & 279k & 165.2k \\
& BookCorpus2 & 43k & 103.6k \\
& PG19 & 20k & 137.4k \\
\midrule[0.5pt]
\multirow{4}{*}{\textbf{Code}} & C & 40k & 108.3k \\
& C++ & 40k & 105.7k \\
& Java & 40k & 94.2k \\
& Python & 40k & 87.9k \\
\bottomrule
\end{tabular}}
\caption{Training datasets overview. \# Instance represents the number of instances, and \# Len represents the average length of each instance in the dataset.}
\label{tab:train-data}
\end{table}

\subsection{Average Long-Dependency Score}
\label{sec: avg lds}
In the experiments, we primarily compare the results of utilizing the entire dataset (Full), randomly selecting a 50\% subset (Rand), and selecting the top-scoring 50\% subset (ProLong) for extending training. We analyzed the average long-dependency score in these three scenarios, as shown in Table \ref{tab:train-data-lds}, where the results for Rand and Full are very close across different data sources, but the results for ProLong are significantly higher than both of them.

\begin{table}[h]
\centering
\resizebox{1.0\columnwidth}{!}{\begin{tabular}{llccc}
\toprule
\multirow{3}{*}{\textbf{Data Type}} & \multirow{3}{*}{\textbf{Datasets}} & \multicolumn{3}{c}{\textbf{Average LDS}}     \\
\cmidrule(lr){3-5}
& & \textbf{Full} & \textbf{Rand} & \textbf{ProLong} \\
& & (100\%) & (50\%) & (50\%) \\
\cmidrule(lr){1-2}\cmidrule(lr){3-5}
\multirow{4}{*}{\textbf{English Book}} & arXiv & 1418.0 & 1418.0 & 1890.3 \\
& Book3 & 385.4 & 386.5 & 1028.9 \\
& BookCorpus2 & 327.2 & 326.2 & 544.1 \\
& PG19 & 469.9 & 467.3 & 1202.9 \\
\midrule[0.5pt]
\multirow{4}{*}{\textbf{Code}} & C & 796.1 & 798.4 & 2484.0 \\
& C++ & 1239.3 & 1240.0 & 2909.6 \\
& Java & 1219.9 & 1219.2 & 2234.1 \\
& Python & 828.7 & 824.2 & 1758.8 \\
\bottomrule
\end{tabular}}
\caption{The average LDS under different data selection strategies on the training datasets.}
\label{tab:train-data-lds}
\end{table}

\section{Experimental Details}
\subsection{Training Hyper-parameters}
\label{sec: hpara} 
All model variants are trained via the next token prediction objective.
We set a learning rate of $2 \times 10^{-5}$ with no weight decay and the whole training step is set to $6000$ with a global batch size of $128$. For subsequent experiments with 50\% data setting, the training step is set to 3000 steps. Additionally, we adopt a linear learning rate warm-up scheduler over $20$ steps and the AdamW \cite{loshchilov2017decoupled} optimizer with $\beta_1 = 0.9$ and $\beta_2 = 0.95$.

\subsection{Model Baselines}
\label{sec: model baseline}
We evaluate several mainstream LLMs with long context capability, including GPT-4-32k~\cite{achiam2023gpt}, GPT-3.5-Turbo-16k \cite{chatgpt}, Llama2-7b \cite{touvron2023llama2}, Llama-13b, Code Llama-7b \cite{roziere2023code}, Yarn-7b-64k \cite{peng2023yarn}, Yarn-13b-64k, Llama2-7B-chat-4k \cite{touvron2023llama2}, Llama2-13B-chat-4k, LongChat-v1.5-7B-32k \cite{longchat2023}, Vicuna-v1.5-7B-16k \cite{vicuna2023}, Vicuna-v1.5-13B-16k, Longlora-7B-16k \cite{chen2023longlora}, and PI-Llama2-13B-16k \cite{chen2023extending}. 

\begin{figure}[h]
    \centering
    \includegraphics[width=1.0\columnwidth]{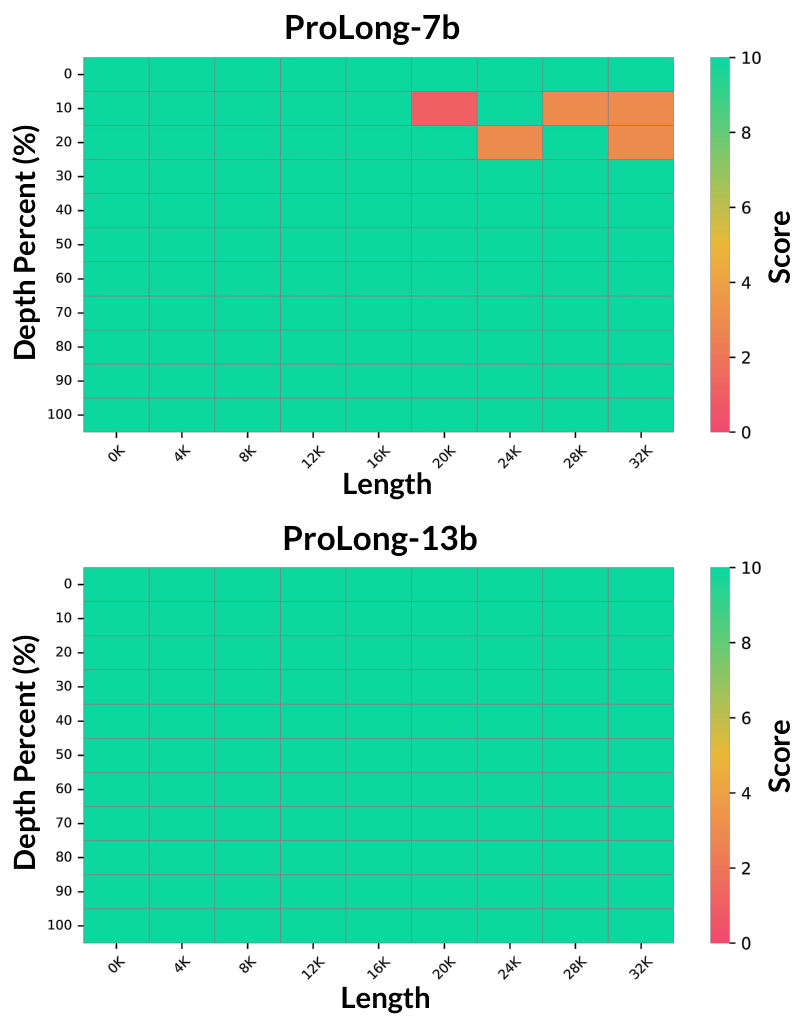}
    \caption{0k-32k pressure test~(Needle In A HayStack) performance of ProLong-7B and ProLong-13B.}
    \label{fig:needle}
\end{figure}

\begin{table*}[t]\normalsize
\centering
\resizebox{0.9\textwidth}{!}{
\begin{tabular}{lcccccc|c}
\toprule
Model       & ARC-c  & Hellaswag & MMLU & TruthfulQA & Winogrande & GSM8K & \textbf{Avg.} \\
\midrule[0.5pt]
Llama2-7b   & 0.52 & 0.79      & 0.46 & 0.39       & 0.69       & 0.13  & 0.50 \\
ProLong-7b  & 0.54 & 0.79      & 0.45 & 0.40       & 0.68       & 0.14  & 0.50 \\
\midrule[0.5pt]
Llama2-13b  & 0.60 & 0.82      & 0.55 & 0.37       & 0.72       & 0.23  & 0.55 \\
ProLong-13b & 0.60 & 0.82      & 0.53 & 0.38       & 0.72       & 0.24  & 0.55 \\
\bottomrule
\end{tabular}}
\caption{Performance on a subset of standard short-length benchmarks.}
\label{tab:short-task}
\end{table*}

\begin{figure*}[t]
    \centering
    \includegraphics[width=0.9\textwidth]{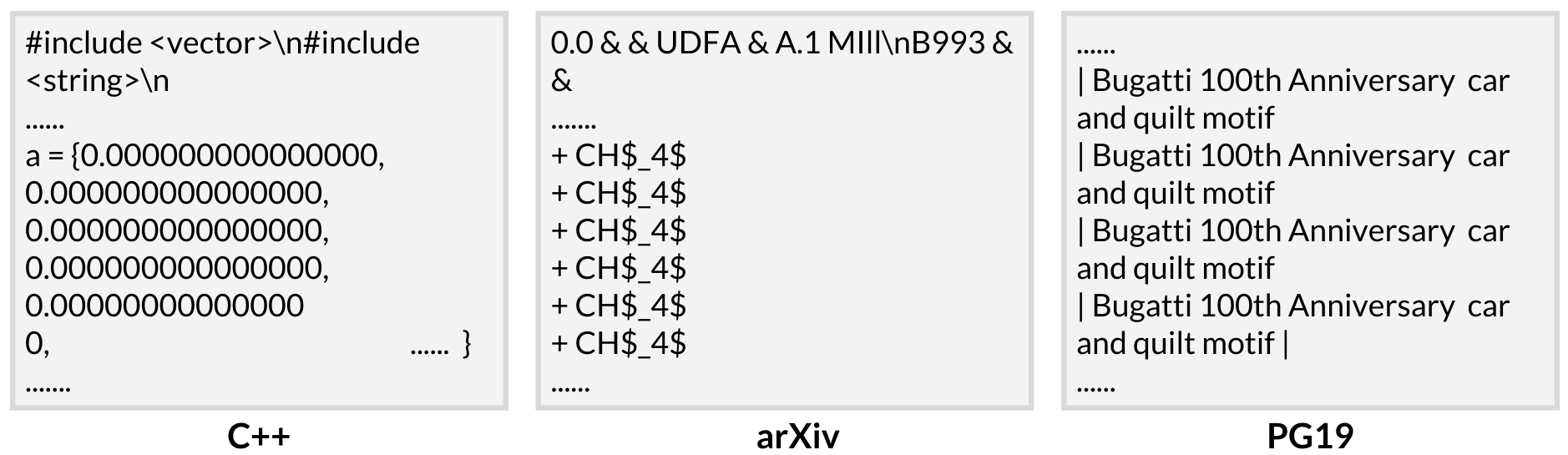}
    \caption{Repetitive patterns examples from arXiv, C++, PG19 datasets.}
    \label{fig:repet}
\end{figure*}

\section{More Experimental Results}
\subsection{Needle in A HayStack Task}
\label{sec: needle}
A simple `needle in a haystack' analysis to test in-context retrieval ability of long context LLMs. 
This test involves placing a random fact or statement (the `needle') in the middle of a long context window (the `haystack'), requiring the model to retrieve this needle. Then iterate over various document depths (where the needle is placed) and context lengths to measure performance.
In the setting where the maximum context length is 32k, the results in Figure \ref{fig:needle} show ProLong-13B obtains 100\% needle in a haystack accuracy across all tested depths and context lengths. ProLong-7B exhibite a few errors when the context length exceeded 16k, it still maintain an overall accuracy of 95\%.

\subsection{Standard short-length benchmarks}
\label{sec: short-bench}
As shown in the table~\ref{tab:short-task}, we evaluate ProLong-7b and ProLong-13b on several standard short-length benchmarks from Hugging Face Open LLM Leaderboard~\cite{open-llm-leaderboard}, i.e., 25-shot ARC-Challenge~\cite{clark2018think}, 10-shot Hellaswag~\cite{zellers2019hellaswag}, 5-shot MMLU~\cite{hendrycks2020measuring}, 0-shot TruthfulQA~\cite{lin2021truthfulqa}, 5-shot Winogrande~\cite{sakaguchi2021winogrande}, 5-shot GSM8K~\cite{cobbe2021training}. When compared with the baseline Llama2-7b and Llama2-13b models, we find that ProLong does not exhibit a significant performance decline, on the contrary, it achieves comparable or even better performance than baseline. 

\section{Additive Dependency Specificity}
\label{sec: add dsp}
Additive Dependency Specificity is calculated as follows:
\begin{equation}
\label{add dsp}
\begin{aligned}
\mathrm{LDS} = \sum_{i = 1}^N \sum_{j = 1}^{i - 1}
& \left\{\right.\left(\right.\alpha \mathrm{DST}_{i, j} + \beta \mathrm{DDI}_{i, j} \\
& + \gamma\mathrm{DSP}_i \left.\right) \times I_{i, j}\left.\right\}
\end{aligned}
\end{equation}
where $\gamma$ is a hyper-parameters. In the ablation experiments, we set the hyper-parameters $\alpha = \beta = \gamma = 1$ in LDS. 

\section{Repetitive Patterns}
\label{sec: repetitive patterns}
As shown in Figure \ref{fig:repet}, we present some examples of repetitive patterns in the training data.
To be specific, the repetitive patterns in C++ are a huge number of reptitive numbers, in the arXiv are numerous repetitive formula symbols, and in the PG19 are an abundance of reptitive tables, which may potentially have a negative impact.
When we do not use dependency specificity or use additive dependency specificity, this type of data will receive a high long-dependency score, thereby impacting the quality of the filtered data.

\end{document}